\documentclass[letterpaper, 10 pt, conference]{ieeeconf}
\IEEEoverridecommandlockouts
\usepackage{amsmath}
\usepackage{cite}
\usepackage{bbold}
\usepackage{epsfig}
\usepackage[caption=false,listofformat=subsimple,labelformat=simple]{subfig}

\usepackage{multirow}
\usepackage{cellspace}
\usepackage{hyperref}
\usepackage{multicol}

\title{\LARGE \bf
MuNES: Multifloor Navigation Including Elevators and Stairs
}
\author{Donghwi Jung, Chan Kim, Jae-Kyung Cho, and Seong-Woo Kim%
\thanks{The authors are with Seoul National University, Seoul, South Korea.\newline
 {\tt\footnotesize \{donghwijung,chan\_kim,jackyoung96,snwoo\}@snu.ac.kr}}}
\begin{document}
\maketitle
\thispagestyle{empty}
\pagestyle{empty}

\begin{abstract}
We propose a scheme called MuNES for single mapping and trajectory planning including elevators and stairs. Optimized multifloor trajectories are important for optimal interfloor movements of robots. However, given two or more options of moving between floors, it is difficult to select the best trajectory because there are no suitable indoor multifloor maps in the existing methods. To solve this problem, MuNES creates a single multifloor map including elevators and stairs by estimating altitude changes based on pressure data. In addition, the proposed method performs floor-based loop detection for faster and more accurate loop closure. The single multifloor map is then voxelized leaving only the parts needed for trajectory planning. An optimal and realistic multifloor trajectory is generated by exploring the voxels using an A* algorithm based on the proposed cost function, which affects realistic factors. We tested this algorithm using data acquired from around a campus and note that a single accurate multifloor map could be created. Furthermore, optimal and realistic multifloor trajectory could be found by selecting the means of motion between floors between elevators and stairs according to factors such as the starting point, ending point, and elevator waiting time. The code and data used in this work are available at \url{https://github.com/donghwijung/MuNES}.
\end{abstract}
\section{Introduction}
For indoor robots to perform their tasks smoothly, movement between floors is essential. With regard to multifloor movement, various optimal trajectory plans between floors, including interfloor movements, have been formulated and studied \cite{yan2021indoor, mavros2022human} on the basis of the traveling salesman problem (TSP). However, most of these studies are theoretical in nature, and practical applications have been studied only for humans. Therefore, this study applies the TSP to indoor multifloor movements of robots. To solve this problem, indoor structures that include the means of movement between floors, such as elevators and stairs, must be represented quantitatively. These representations can also be applied to localization of the robot. Therefore, multifloor indoor maps, including elevators and stairs, are required for autonomous multifloor navigation.\\
\indent Two main approaches are used to create the existing multifloor maps. The first method involves creating a separate map for each floor \cite{ozkil2011mapping,karg2010consistent}, and the second method involves creating a single map with connected floors, including movements between the floors \cite{zhang2014loam, wei2021ground}. In the former approach, some floors have complex structures that make mapping easy, while others may have shorter maps owing to simple structures, such as long featureless corridors. In the latter approach, when the sensor data acquired by LiDAR or camera change according to the movement of the robot, such as for stairs, the data can be detected and represented on a map. On the other hand, when the sensor data are constant even though the robot location changes, such as inside an elevator, it is difficult to detect movement and hence create a map. To solve these problems, the following methods are used. First, air pressure changes are measured in all indoor sections, including inside the elevator, and used as a constraint on altitude changes; this constraint detects altitude changes inside the elevator and prevents the predicted pose graph from being distorted before and after movement between floors. In addition, multifloor navigation including elevators and stairs (MuNES) uses floor-based loop detection to find loops quickly and more accurately than without floor labels. The loops found in this manner are utilized to minimize the odometry estimation errors that can occur because of movement between floors.\\
\begin{figure}[t]
    \centering
    \framebox{\parbox{0.4\textwidth}{\includegraphics[width=0.4\textwidth]{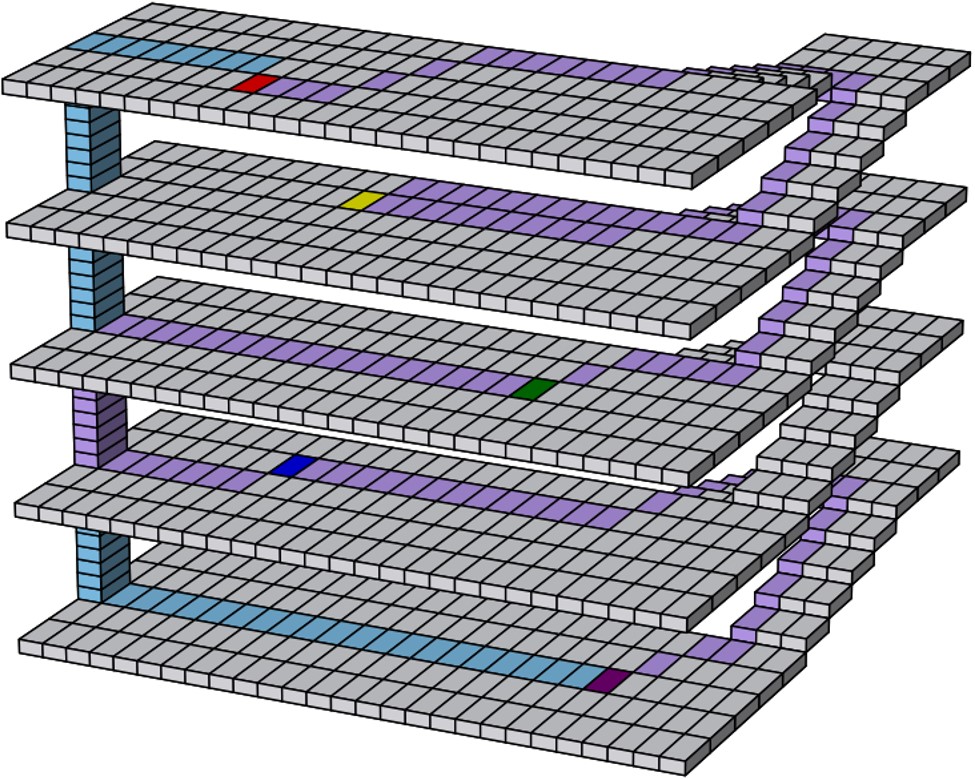}}}
    \caption{Result of multidestination trajectory planning including intermediate destinations. The purple and blue colors indicate trajectory between destinations and returning trajectory to the starting point, respectively. The other colors from red to dark purple represent the starting, ending, and intermediate pints. The order of points for the trajectory is red, yellow, green, blue, and dark purple.}
    \label{fig:multi_f}
\end{figure}
\begin{figure*}[t]
    \centering
    \framebox{\parbox{\textwidth}{\includegraphics[width=\textwidth]{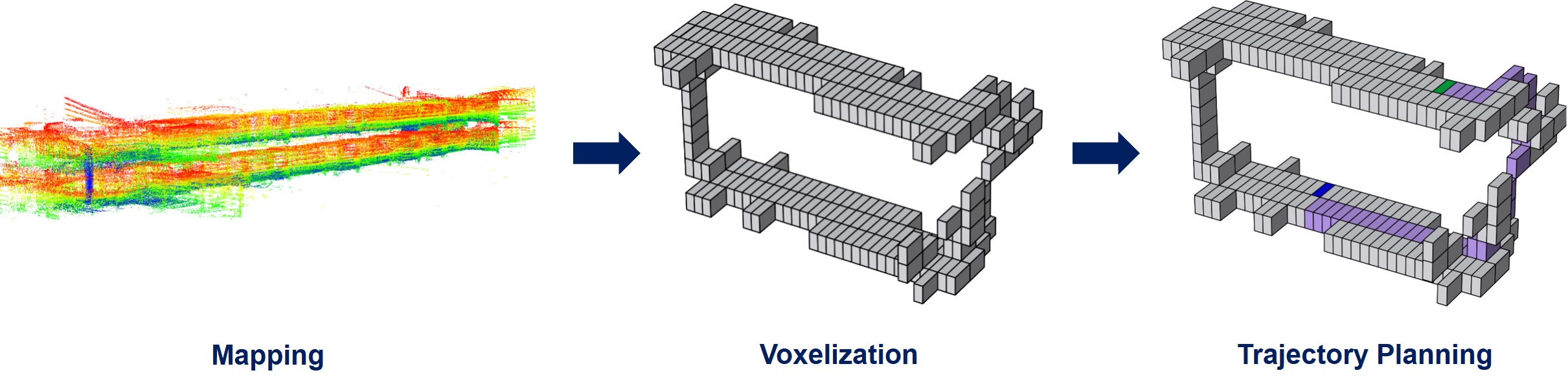}}}
    \caption{Process of MuNES; the point cloud map (left) is voxelized and represented as a voxel map (middle); MuNES uses this voxel map to proceed with multifloor trajectory planning (right).}
    \label{fig:process}
\end{figure*}
\indent Multifloor trajectories are generated based on multifloor maps. Existing studies on multifloor trajectory planning are connected with two methods: creating a separate trajectory for each floor and later connecting them \cite{lee2016multi}; creating a trajectory that includes movement between floors \cite{baltashov2018path}. In the former case, trajectories from the elevator or stairs to the starting and ending points are created for the departure and destination floors, respectively, and these are connected to obtain a single trajectory. In this method, the trajectories inside stairs or elevators are not considered. Therefore, it is difficult to accurately compare trajectories; hence, there is a possibility that the actual final trajectory is not optimal. In the latter case, some existing studies have used stairs to move between floors, but these studies do not consider trajectories inside elevators. Because elevators are installed in many buildings and many robots use such elevators to move between floors, it is necessary to plan a trajectory that includes the elevators for better usability. Therefore, the proposed trajectory planning uses a map that includes both elevators and stairs. In addition, it is possible to find optimal and realistic trajectories by formulating the cost function of the A* algorithm \cite{hart1968formal} by considering factors such as the waiting time for the elevator.\\
\indent To the best of our knowledge, MuNES is the first application of the multifloor TSP (MF-TSP) described in \cite{yan2021indoor, mavros2022human} to navigation of a mobile robot. The performance of MuNES was confirmed by tests on data from various situations around an university campus, such as those with both elevators as well as stairs and with only elevators. As a result, MuNES is shown to solve the MF-TSP of mobile robots.\\
\indent The main contributions of this work are as follows:
\begin{itemize}
    \item \emph{Application of MF-TSP to Mobile Robots}: We first apply the MF-TSP to a mobile robot. MuNES finds the optimal multifloor trajectory that the robot can follow.
    \item \emph{Accurate Multifloor Mapping}: Using floor-based loop detection and odometry correction with pressure data, we build a single multifloor map with greater accuracy than conventional methods.
    \item \emph{Realistic Multifloor Trajectory}: When planning a multifloor trajectory, a realistic trajectory is created by considering factors such as waiting time for the elevator.
\end{itemize}
\section{Related Works}
\subsection{Mapping}\label{rel_works_mapping}
Ozkil \emph{et al.} \cite{ozkil2011mapping} mapped each floor in 2D space using the method proposed by Grisetti \emph{et al.} \cite{grisetti2007improved} and aligned each map using pressure data. The pressure data were used to segment the map of each floor. In another study \cite{fan20173d}, after changing the depth image to a point cloud, each floor was mapped in 3D space; after calibrating the pressure data obtained from each floor using an extended Kalman filter, the elevation was estimated using the same method as in \cite{ozkil2011mapping}. In studies \cite{ozkil2011mapping} and \cite{fan20173d}, the estimated elevations were used only to align the maps of each of the floors and such alignments were not related to correction of errors that occurred during mapping. On the other hand, some researchers have corrected relatively inaccurate maps based on the most accurate among several 2D maps representing each floor. As in study \cite{karg2010consistent}, 2D maps of each floor were created individually, and the relative constraints obtained through Monte Carlo localization were applied for optimization. This optimization was able to correct the maps of the inaccurately mapped floors owing to their simple structures. However, the means of moving between the floors, such as elevators and stairs, were not included in these maps.\\
\indent LiDAR odometry and mapping (LOAM) \cite{zhang2014loam} is a representative algorithm that is used to create a single multifloor map using LiDAR as the main sensor. This method can be used for mapping when the surrounding features change with movement, such as stairs, however not when the surrounding features remain constant regardless of position change of robots, such as inside an elevator. The disadvantages of LOAM could be supplemented using additional sensors, such as inertial measurement units (IMUs) \cite{shan2018lego,qin2020lins,shan2020lio,kang20223d}. The position changes over time are determined by measuring the acceleration of the robot through the IMU and integrating it. In cases involving elevators, acceleration in the $z$ direction occurs as the elevator ascends and descends. Thus, changes in the elevation are detected. However, in the case of prediction using IMUs, the lengthier the interfloor movement, the more are accumulations of the prediction errors. Therefore, in this study, instead of the IMU, we exploit the pressure data, which are directly connected to the absolute elevation, to estimate the elevation reliably regardless of the degree of movement between floors. Another approach to create a single multifloor map through LiDAR is ground simultaneous localization and mapping (ground SLAM) \cite{wei2021ground}. In ground SLAM, the ground was used as a landmark after ground extraction and applied as a constraint to the robot pose. This method is effective for scenarios involving flat floors and slopes. However, if the floor is uneven, as in the case of stairs, or if there are no changes between the robot pose and floor, as in the case of elevators, the algorithm is ineffective.
\subsection{Trajectory Planning}
Existing multifloor trajectory planning methods are of two types, namely those that consider the movement trajectory between floors and those that do not. First, if the interfloor movement is not considered, then trajectory planning on the floor containing the starting and ending points is carried out separately, and the corresponding trajectories are connected to form the complete trajectory \cite{lee2016multi}. This method does not consider the trajectory of movement between floors through stairs or elevators. Therefore, when there are several choices of movement between floors, it is impossible to find an optimal trajectory because these choices cannot be compared. There is an existing study that considers a trajectory involving stairs in the trajectory planning \cite{baltashov2018path}; in this case, the trajectory is planned from the floor with the starting and ending points to the stairs, and combined with the trajectory inside the stairs to compose the complete trajectory. Such trajectory planning has the advantage of producing a trajectory close to reality by configuring the entire trajectory in consideration of the trajectory between floors. However, elevator were not included in \cite{baltashov2018path}; because many robots use elevators to move between floors, this approach is difficult to apply in various multifloor situations except for elevators.\\
\indent Considering these previous works, there is no single appropriate method for applying MF-TSP to mobile robot scenarios. The following sections therefore describe the MuNES proposed to solve this problem in detail.
\begin{figure}[t]
    \centering
    \framebox{\parbox{0.48\textwidth}{\includegraphics[width=0.48\textwidth]{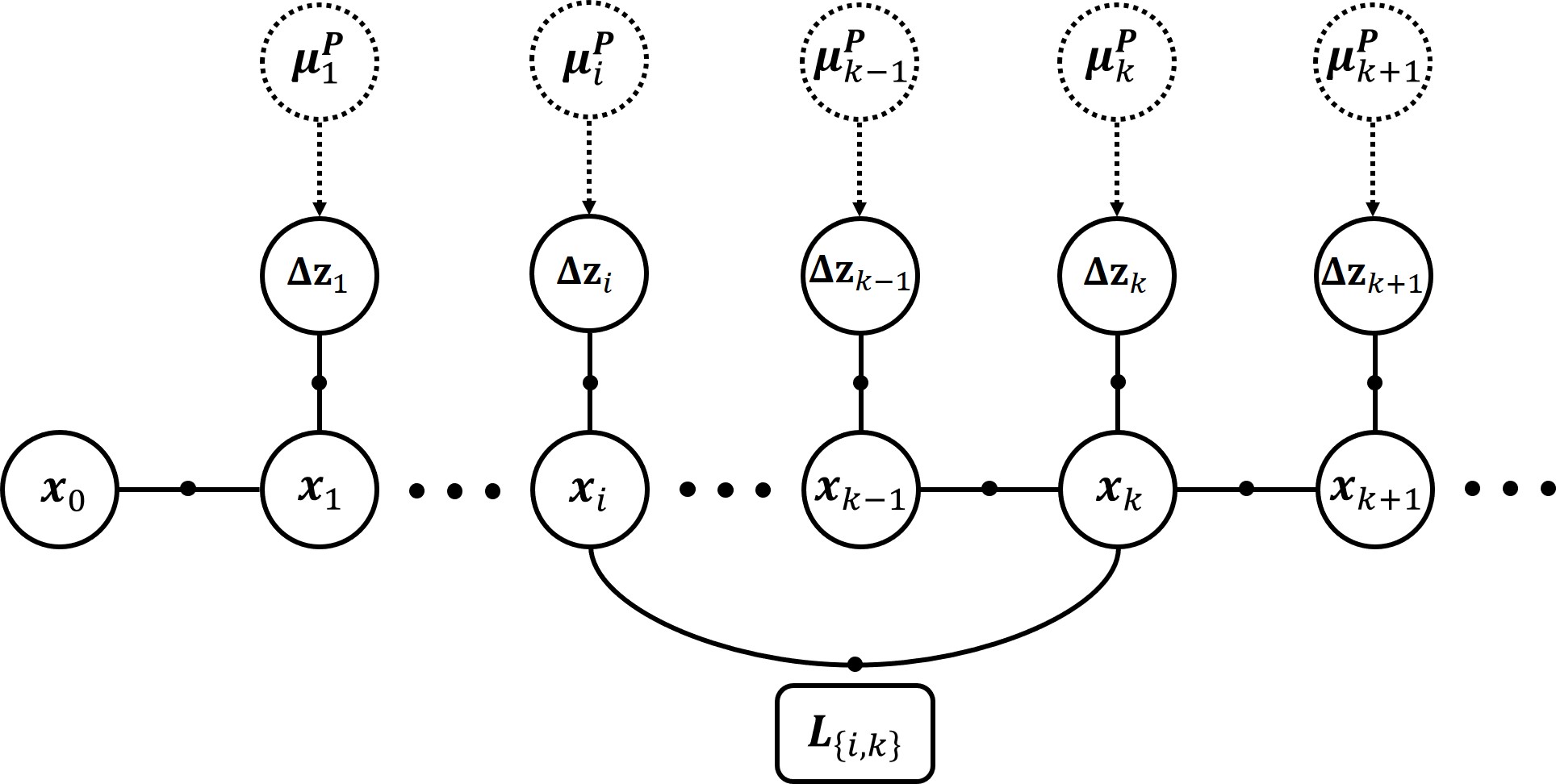}}}
    \caption{System architecture of the mapping part in MuNES.}
    \label{fig:architecture}
\end{figure}
\section{Methods}
The process of MuNES is represented in Fig. \ref{fig:process} and consists of three parts: mapping, voxelization, and trajectory planning.
\subsection{Mapping}\label{mapping}
The basic structure for mapping is shown in Fig. \ref{fig:architecture}. This framework involves elevation estimation, elevator detection, and floor-based loop detection.
First, the change in the elevations of an elevator are estimated as follows:
\begin{equation}\label{eqn:elevation_estimation}
    \Delta{z_i} = 44330\cdot \left(1-\left(\frac{\mu^P_i}{P_{cri}}\right)^{\frac{1}{5.255}}\right),
\end{equation}
where $\mu^P_i$ represents the average of the sequential pressure data accounting for the pressure fluctuations. $P_{cri}$ is the reference pressure value measured at the starting position and Eq. \eqref{eqn:elevation_estimation} is derived from the international pressure equation \cite{milette2012professional}. Furthermore, $\Delta{z_i}$ denotes the change in $z$ of the $i$th robot pose, and these $\Delta z$ are added to the graph as elevation constraints. These constraints can refine the $z$ values for each pose as a constraint during graph optimization. In addition, by comparing the predefined threshold and $\Delta z$, MuNES estimates the change of current floor. If the absolute value of $\Delta z$ is greater than the threshold, $1$ is added or subtracted from the current floor depending on the sign of $\Delta z$. This estimated change of floor is used in loop detection as a label.\\
\indent To include an elevator in the map, it is necessary to check whether the robot is inside the elevator or not. For this purpose, MuNES exploits the average of the squares of the distances of the points in the point cloud because the point cloud is distributed relatively close to the robot inside the elevator. If the distance is less than the predefined threshold, MuNES determines that the robot is inside the elevator. Moreover, if it is determined that the robot is inside the elevator, the distribution of the acquired point cloud is arbitrarily changed to the shape of a hollow cuboid. In other words, the point cloud is stacked in the $z$-axis direction using the elevation changes estimated through Eq. \eqref{eqn:elevation_estimation}.\\
\indent To perform loop detection, \emph{Scan Context}\cite{kim2018scan} is used. To reduce the number of false positives detected in an indoor environment with a similar structure, the obtained current floor is used for labeling. The performance of loop detection is enhanced by performing detection in comparison with the titles with the same labels. By comparing contexts, $L_{\{i,k\}}$ described in Fig. \ref{fig:architecture}, which indicates the loop the between the $i$th and $k$th nodes is detected. Based on the basic structure in \cite{kim2018scan}, MuNES creates ring key trees with floor labels. Therefore, the tree of the corresponding floor is searched using the floor label to find the loop candidate. The detected loops are then attached to the graph as constraints. Subsequently, the graph is optimized using incremental smoothing and mapping with the Bayes tree (iSAM2) \cite{kaess2012isam2}.
\subsection{Voxelization}\label{voxelization}
For direct use in trajectory planning, MuNES voxelizes the multifloor map, leaving only the bottoms of corridors, stairs and elevators of each floor. A multifloor indoor map used for trajectory planning is shown in the middle of Fig. \ref{fig:process}. The bottom surface of each floor is represented by a 2D array of voxels. Each floor is connected by stairs and elevators, and the stairs are represented by diagonal voxels. In addition, considering that a robot can hardly move in the $x,y$ direction inside an elevator, the voxels are represented as a stack in the $z$ direction. For voxelization, only the floor points of the corridor are extracted first. In this case, MuNES utilizes the channel information of the LiDAR. By selecting a point containing a channel lower than the predefined maximum ground channel, MuNES chooses a ground point based on the assumption that the LiDAR is installed parallel to the ground. In the case of voxel representations of stairs and elevators, only the top $N_z$ voxels containing relatively many points based on the same $z$ value among the remaining voxels except for the ground, are retained, and the rest are removed. These voxels denote the free space in which the robot can move. Therefore, when planning a trajectory, the algorithm searches the voxels to find the optimal trajectory.
\begin{figure}[t]
    \centering
    \framebox{\parbox{0.45\textwidth}{\includegraphics[width=0.45\textwidth]{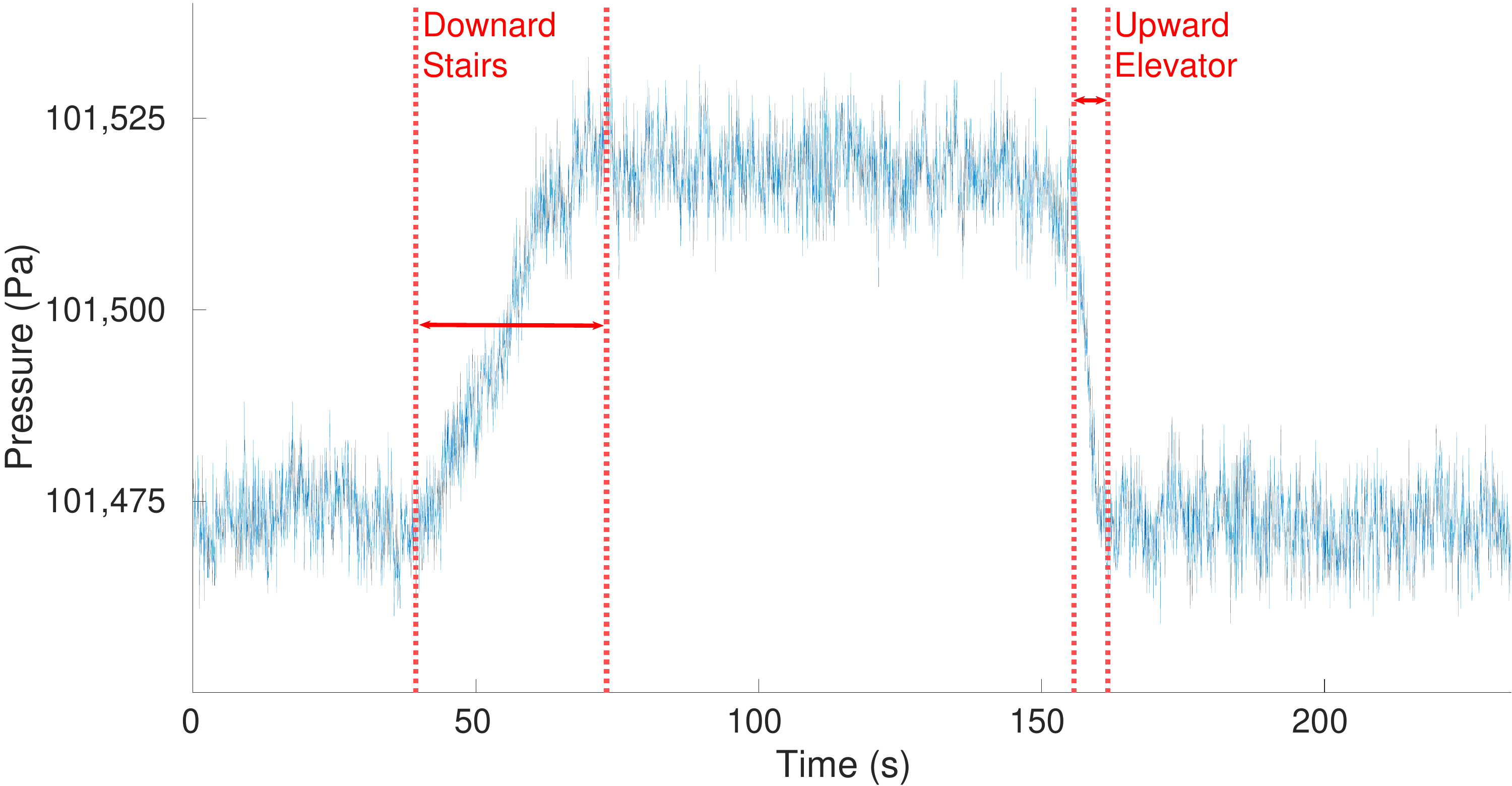}}}
    \caption{Pressure change from the third to the second floor in building $1$.}
    \label{fig:pressure_graph}
\end{figure}
\begin{figure*}
    \begin{multicols}{2}
    \hfill
    \subfloat[A map created at an angle after moving between floors through stairs.]{
	   \framebox{\parbox{0.55\textwidth}{\includegraphics[width=0.55\textwidth]{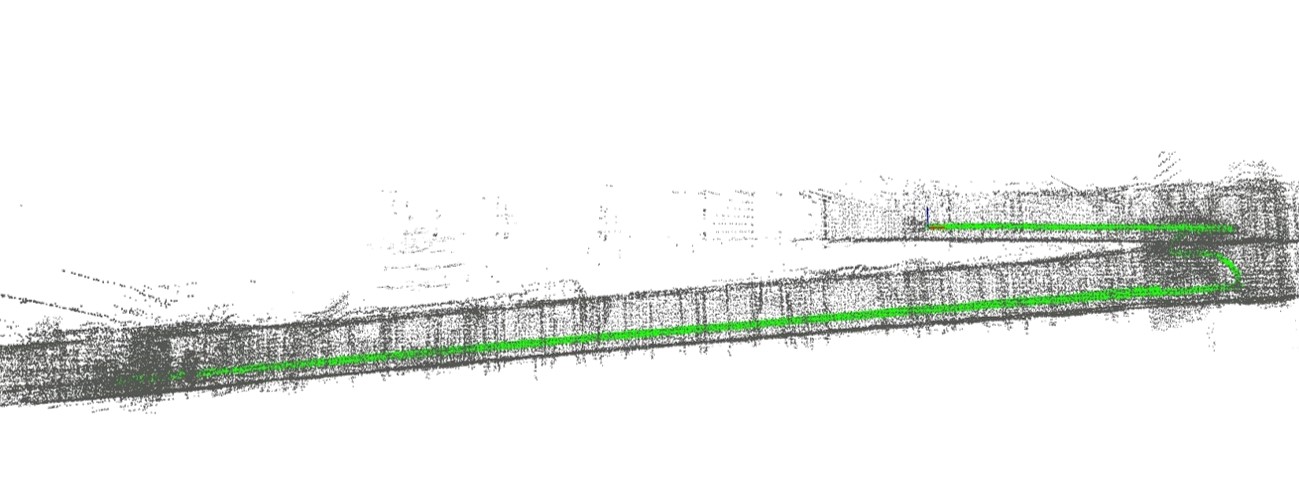}}}
        \label{fig:loam}
    }
    \hfill
    \subfloat[A well-ordered map created with pressure constraints applied.]{
  	   \framebox{\parbox{0.55\textwidth}{\includegraphics[width=0.55\textwidth]{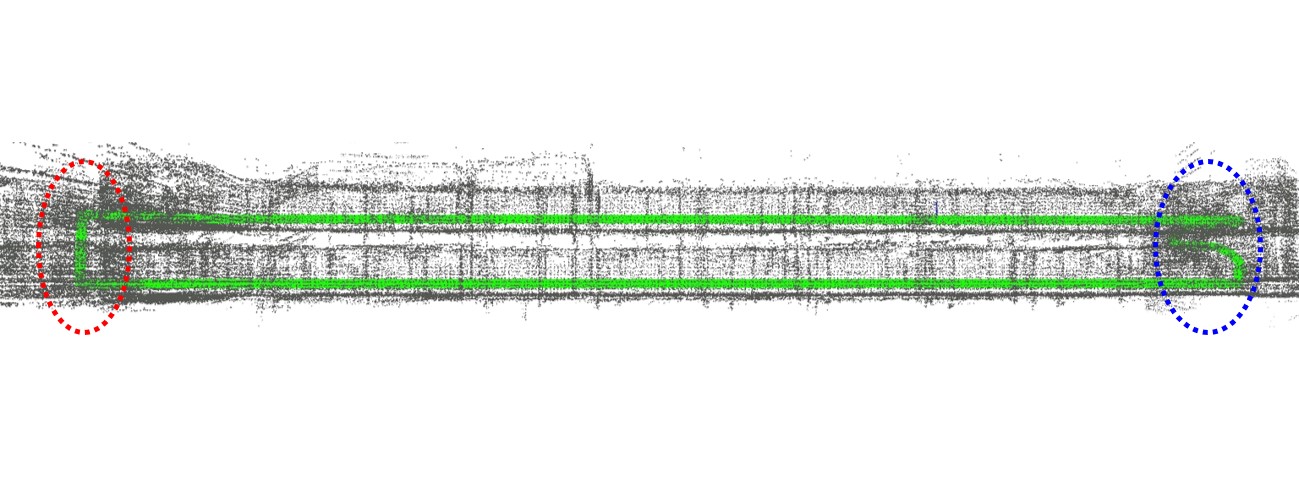}}}
        \label{fig:mufe}
    }
    \hfill
    
    \hfill
    \subfloat[An overlapped map of two floors without an elevator.]{
  	   \framebox{\parbox{0.4\textwidth}{\includegraphics[width=0.4\textwidth]{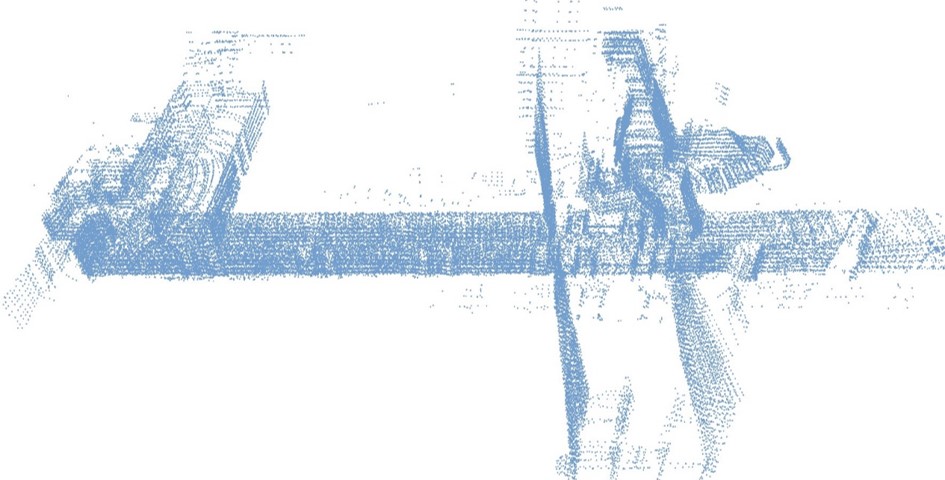}}}
        \label{fig:loam2}
    }
    \hfill
    
    \hfill
    \subfloat[A separate map of two floors with elevators represented.]{
 	   \framebox{\parbox{0.4\textwidth}{\includegraphics[width=0.4\textwidth]{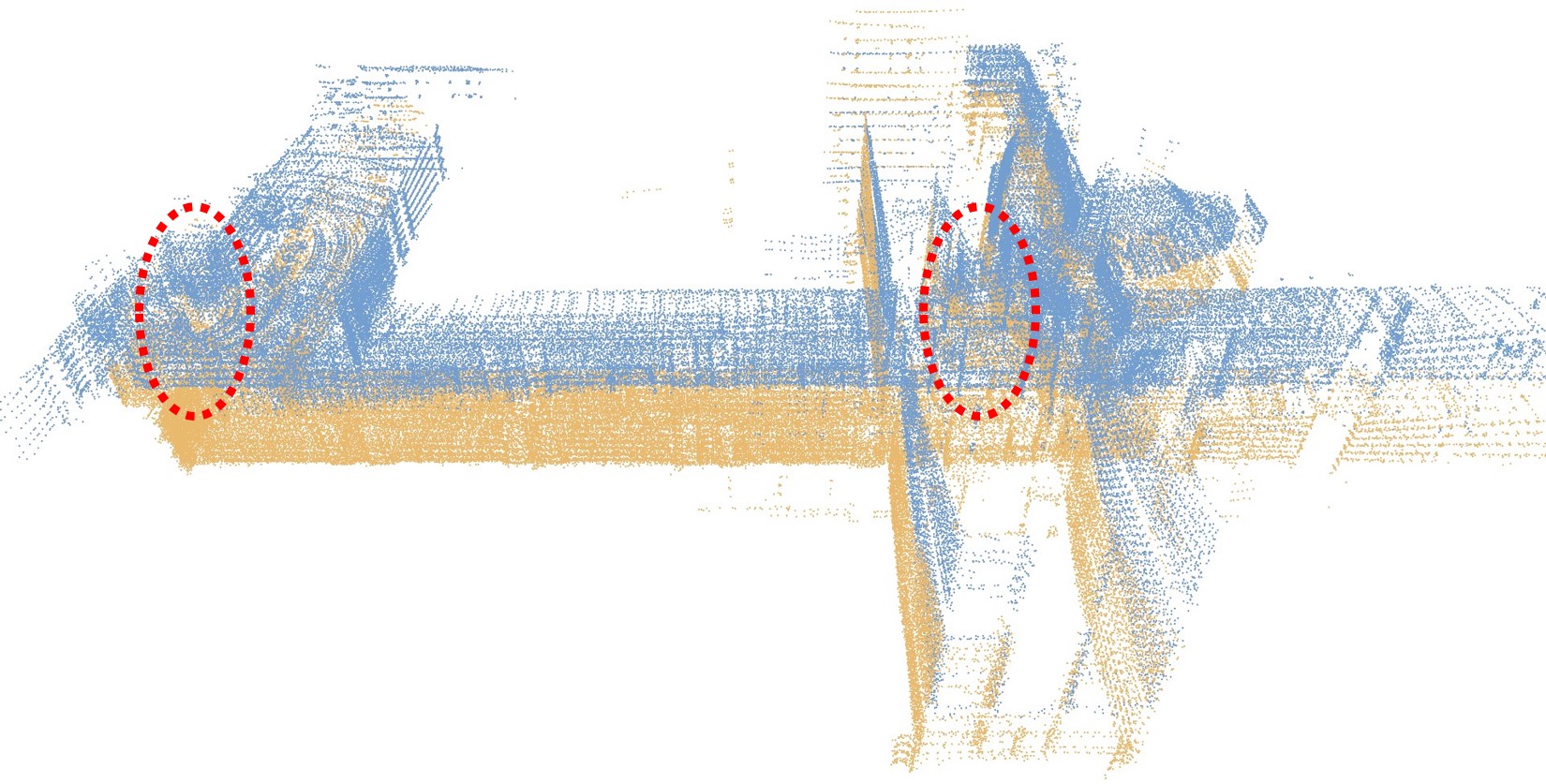}}}
        \label{fig:mufe2}
    }
    \end{multicols}
    \vspace{-0.4cm}
    \caption[]{\subref{fig:loam},\subref{fig:loam2} Results of multifloor mapping using LOAM and \subref{fig:mufe}, \subref{fig:mufe2} the proposed method; \subref{fig:loam}, \subref{fig:mufe} denote the results of mapping with an elevator (red) and stairs (blue), and \subref{fig:loam2}, \subref{fig:mufe2} represent the results of mapping with elevators on both sides (red).}
\end{figure*}
\subsection{Trajectory Planning}\label{trajectory_planning}
Before adding the corresponding vertex to a trajectory, MuNES judges whether this vertex is accessible. First, MuNES checks whether the node to be added to the trajectory is included in $S$, which is a set of voxels acquired as shown in Sec. \ref{voxelization}. Additionally, if any node among the neighboring voxels between a robot's current position $p^{crt}$ and target position $p^{tgt}$ is not included in $S$, the trajectory is determined to be infeasible owing to collision with a boundary, such as a wall. Finally, the remaining cases are judged as normal voxels and added to the trajectory. If it is determined that a node is accessible, the cost function is calculated for that node. Here, A* algorithm\cite{hart1968formal} is used to create the trajectory. In the case of the cost function, to plan a trajectory that includes more realistic movement between floors, the waiting time for the elevator, speed of the elevator, and speed of the robot are considered. The cost function is then formulated as follows:
\begin{gather}
    f(n)=g(n)+h(n),\\
        h(n)=
    \begin{cases}
	    h_1(n), & \text{if elevators},\\
        h_2(n), & \text{else},
	 \end{cases}\\
 t^{elv}_{wtg}=\frac{|p^{elv}.z-p^{str}.z|}{V^{elv}},\;\;\;t^{elv}_{opr}=\frac{|p^{tgt}.z-p^{crt}.z|}{V^{elv}},\label{t_elv}\\
    h_1(n)=t^{elv}_{wtg} + t^{elv}_{opr},\;\;\;
    h_2(n)=\frac{\|p^{tgt}-p^{crt}\|_2}{V^{rbt}},\label{h_1_and_2}
\end{gather}
where $n$ represents the current node, $f$ denotes the evaluation function, and $g$ indicates the cost of the trajectory from the starting to $n$th nodes. Additionally, $h$ denotes the estimated cost of trajectory from node $n$ to the target node. Moreover, $V^{rbt}$ and $V^{elv}$ are the speeds of the robot and elevator, respectively. In this case, we assume that the elevator moves at a constant speed, $V^{elv}$. Furthermore, $p^{elv}$ and $p^{str}$ are the current position of the elevator and position of the starting point. In addition, the $z$ value for each $p$ is represented as $p.z$, and $t^{elv}_{wtg}$ denotes the elevator waiting time, which is the time required for the elevator to reach the robot's current floor. Additionally, $t^{elv}_{opr}$ is the operating time of the elevator, which is the time taken from the departure to the destination floors on the elevator. Because the elevator only moves along the $z$ axis, the values of $x,y$ are not considered in Eq. \eqref{t_elv}. Furthermore, as in Eq. \eqref{h_1_and_2}, $h_1$ is same as the sum of $t^{elv}_{wtg}$ and $t^{elv}_{opr}$, which is the total time spent waiting for and boarding the elevator. In other cases, including stairs, $h_2$ is calculated as the estimated time from the current to target points by the movement of the robot.\\
\section{Experiments}
\subsection{Data Collection and Implementation}
Data collection was performed using \emph{Velodyne VLP 16} as the LiDAR, \emph{Xsens MTi 680G} as the IMU, and a barometer. In addition, we used $100$ sequential pressure data to calculate $\mu_P$ in Eq. \eqref{eqn:elevation_estimation}. Moreover, $2.5\,m$ was applied as the threshold for elevation estimation, and the maximum ground channel was set to $4$. In addition, $N_z$ was set to $5$ and $V^{elv}$ and $V^{rbt}$ were both set to $1\,m/s$. The corresponding values were set based on the acquired data.\\
\subsection{Comparisons}
The proposed approach was compared with the LOAM\cite{zhang2014loam} family of methods. MuNES uses 3D LiDAR as the main sensor in the mapping process in the same way as LOAMs, thus we compared these methods with MuNES. In addition, LeGO-LOAM\cite{shan2018lego}, LINS\cite{qin2020lins}, and LIO-SAM\cite{shan2020lio} use IMUs as well as LiDARs to estimate the changes in elevation even in situations where there are no changes in the point cloud over time, such as inside the elevator. Furthermore, optimization using iSAM2\cite{kaess2012isam2} was applied to all methods for comparison. For the metrics, the mean absolute error was used as the error between the ground truth and estimated value. This error is represented by $\mu_{err}$ and $\sigma_{err}$ which are the mean and standard deviation of the mean absolute error, respectively. Moreover, the values for cases involving stairs and elevators were calculated from $10$ repeated trials for each algorithm. In the case of the same floor, $\mu_{err}$ and $\sigma_{err}$ were determined after extracting the $z$ values for all poses within the floor. Additionally, we compared the performances of loop closing with and without floor labels. The results of trajectory planning were then compared by changing the current position of the elevator, which is directly related to the waiting time for the elevator.\\
\begin{table*}[t]
\small
\caption{Accuracy Of Elevation Estimation in the case of Stairs, Elevators, and within the Same Floor.}
\label{table:1}
\begin{center}
    \begin{tabular}{Sc Sc Sc Sc Sc Sc Sc Sc Sc}\hline
    && LOAM\cite{zhang2014loam} & LeGO (LiDAR)\cite{shan2018lego}& LeGO (LiDAR+IMU) & LINS\cite{qin2020lins} & LIO-SAM\cite{shan2020lio} & MuNES \\\hline
    \multirow{2}{*}{Stairs}&$\mu_{err}$ (m)& 0.49& 3.05&  2.91&2.07&-&\textbf{0.36}\\
    &$\sigma_{err}$ &  0.08& 0.63& 0.73&  0.25&-& \textbf{0.03}\\\hline
    \multirow{2}{*}{Elevator}&$\mu_{err}$ (m)&-&-&2.03&3.19&1.44&\textbf{0.53}\\
    &$\sigma_{err}$ &-&-&0.17&0.02&0.90& \textbf{0.14}\\\hline
    \multirow{2}{*}{Same Floor}&$\mu_{err}$ (m)& 0.60& 0.80& - &  1.53&0.82&\textbf{0.26}\\
    &$\sigma_{err}$& 0.49 & 0.61& - &  2.16&1.25&\textbf{0.10}\\\hline
    \end{tabular}
    \vspace{-0.4cm}
\end{center}
\end{table*}
\subsection{Evaluations}
\subsubsection{Mapping}
Figure \ref{fig:pressure_graph} shows the pressure change in the situation involving multifloor movement. The graph exhibits increasing and decreasing trends as the floor varies. The increase in pressure between $39.40$ and $73.12\,s$ corresponds to movement down the stairs, and the decrease in pressure between $155.72$ and $161.68\,s$ is related to upward movement through the elevator. The distinct changes in pressure indicate that the floor transition can be accurately reflected using the barometric approach.\\
\indent In Table \ref{table:1}, loop closing was not used in all methods. If the performance was significantly poor relative to other results, then that result is not listed. The top two rows of Table \ref{table:1} present comparisons of the elevation estimation accuracies for the stair and elevator cases. The ground truth data is $3.64\,m$, which is the distance between two floors in building $1$. MuNES achieves the best estimate for elevation in the case involving stairs, with the least mean and standard deviation of error. This observation indicates that the pressure data can act as a constraint to detect changes in elevation. Moreover, the results of the estimations are highly stable in MuNES. On the other hand, except for LOAM, the prediction accuracies for elevation changes are low in the remaining methods. This result is attributed to shaking in all directions while moving along the stairs. Because the values of the acceleration and angular velocity change continuously within a short period, the prediction accuracy using the inertial sensor is low, which likely affects the overall elevation prediction. Moreover, LeGO-LOAM is tailored to a situation in which a ground vehicle moves along an even floor. Therefore, this algorithm was not suitable for movement on an uneven surface, such as stairs.\\
\indent To estimate the elevations of the elevators, IMU data were used for all methods except MuNES, and barometric data were applied to MuNES. As mentioned in Sec. \ref{rel_works_mapping}, LOAM and LeGO-LOAM with LiDAR only considered that the robot was stationary inside the elevator because there were no changes in their point cloud features over time. Therefore, these algorithms were excluded from the comparison. Although elevation changes were detected by the three methods with IMUs, notable errors against the ground truth were observed. In addition, LIO-SAM achieved a value ($2.83\,m$) close to the ground truth; however, in certain trials, a value ($0.78\,m$) significantly different from the ground truth was obtained, resulting in large $\mu_{err}$ and $\sigma_{err}$. In the case of MuNES, relatively small $\mu_{err}$ and $\sigma_{err}$ were obtained. Therefore, we conclude that MuNES predicts the elevation change inside the elevator most accurately and reliably.\\
\indent The bottom row of Table \ref{table:1} presents a comparison of the elevation estimation accuracy for a given floor. We acquired data by moving along a flat surface without any transition between floors. Therefore, the elevation prediction result was expected to be close to zero. The elevation prediction accuracy was compared with the mean and standard deviation of the $z$ values of the robot poses calculated for that floor. Among the tested methods, the proposed method achieved a mean value closest to zero and the lowest standard deviation. Therefore, we conclude that the pressure data acted as a constraint to maintain the elevation at a constant value in a given floor. In general, the methods using IMU data exhibited inferior performances, because even though there were no changes in altitude when the robot moved without changing floors, the acceleration and angular velocity were measured in the process, and these values were gradually accumulated to increase the prediction errors. In the case of LeGO-LOAM based on only LiDAR, the prediction accuracy was high, unlike the results shown in Table \ref{table:1}. This is because the algorithm was designed to consider movement along an even ground. Table \ref{table:3} presents a comparison of the loop detection performances with and without floor labeling. The computation time for the case without floor labeling was $1.19$ times higher than that with labeling because a larger number of candidates were required for comparison. Furthermore, the precision was higher in the case with floor labeling because of the decrease in the number of false positives.\\
\indent Figures \ref{fig:loam} and \ref{fig:loam2} show the results of LOAM \cite{zhang2014loam} based on multifloor navigation data. Moreover, Figures \ref{fig:mufe} and \ref{fig:mufe2} represent the results obtained using the proposed method. The lower part of the map in Fig. \ref{fig:loam} is not parallel to the upper part. This inclination occurred after passing the stairs. However, as shown in Fig. \ref{fig:mufe}, the slope was eliminated, and each floor of the map was parallel. This improvement was achieved by applying the pressure constraint to each robot pose. In addition, Fig. \ref{fig:loam2} does not indicate an elevator because LOAM did not consider the elevators. In contrast, as seen in Fig. \ref{fig:mufe2}, the multifloor map obtained using the proposed method contains two elevators. This inclusion could be achieved by estimating the elevation changes inside the elevator, as explained in Sec. \ref{mapping}.
\begin{table}[t]
\small
\caption{Loop Detection Results.}
\label{table:3}
\begin{center}
    \resizebox{0.48\textwidth}{!}{
        \begin{tabular}{Sc Sc Sc}\hline
        & Without floor label & With floor label\\\hline
        Computing Time (ms) & 1.27 & \textbf{1.07}\\
        Precision& 0.24&\textbf{0.60}\\\hline
        \end{tabular}
    }
\end{center}
\end{table}
\subsubsection{Trajectory Planning}
\begin{figure}[t]
    \centering
    \framebox{\parbox{0.5\textwidth}{\includegraphics[width=0.5\textwidth]{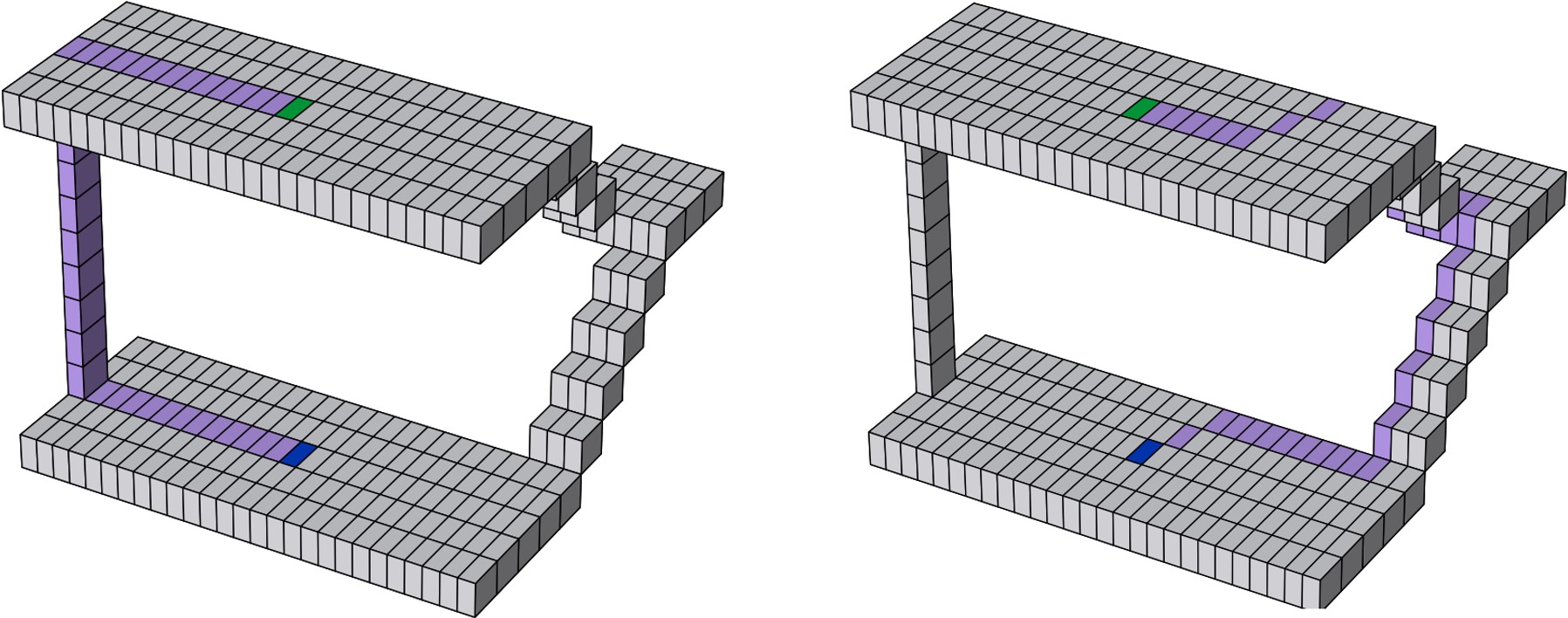}}}
    \caption{Results of different trajectory planning methods depending on variation of waiting time for the elevator. The green, blue, and purple colors indicate the starting point, ending point, and trajectory, respectively.}
    \label{fig:diff_waiting_time}
\end{figure}

Two experiments were performed in this study. First, as shown in Fig. \ref{fig:diff_waiting_time}, we compared the results of trajectory planning by changing the waiting time for the elevator. In the experiment, the waiting time varied depending on the initial position of the elevator. As shown in the left of Fig. \ref{fig:diff_waiting_time}, if the initial floor of the elevator was $1$ which is same as the lower floor, the trajectory through the elevator was chosen. On the other hand, if the initial floor of the elevator was $2$ corresponding to the upper floor, the trajectory through the stairs was selected as represented in the right of Fig. \ref{fig:diff_waiting_time}. Second, we experimented with multidestination trajectory planning using intermediate destinations located from the $5$th to $1$st floors, as represented in Fig. \ref{fig:multi_f}. In this case, for the purpose of the trajectory planning experiment, the voxel map was arbitrarily configured to include stairs and elevators across the five floors. As a result, multidestination trajectory planning utilized different stairs or elevators depending on the locations of the starting and ending points. For the trajectory from the $5$th to $4$th floors, because we set the initial location of the elevator as the $1$st floor which was far from the robot location, the robot selected the stairs instead of the elevator to reach the destination on the $4$th floor. Moreover, when the robot returned from the $1$st to $5$th floors, MuNES selected the elevator instead of the stairs; this is because of the longer distance of movement through the stairs than the elevator. With these results, we conclude that the robot is able to move efficiently by comparing more than one means of movement between floors with MuNES.\\
\section{Conclusion}
This paper proposes the MuNES scheme for mapping and single-trajectory planning including elevators and stairs. The proposed algorithm estimates the altitude using pressure data in all indoor sections, including the inside of the elevator, and uses them as a constraint for odometry estimations. Through this estimation, a change in altitude inside the elevator is detected, and the predicted pose graph before and after movement between floors through the stairs are not distorted. Moreover, by utilizing the floor-based loop detection, faster and more accurate loop closure is possible. Multifloor trajectory planning is performed by applying the A* algorithm composed of the proposed cost function, which reflects realistic factors such as elevator waiting time using the generated multifloor map. By testing this algorithm with data collected around a university campus, an accurate single multifloor map could be created, and an optimal realistic multifloor trajectory could be obtained. This trajectory was planned by selecting the optimal choice of movement between floors, i.e., elevator or stairs, depending on the locations of the starting and ending points as well as the waiting time for the elevator. Thus, we confirm that the MF-TSP can be applied to robot scenarios.\\
\section*{Acknowledgement}
This work was supported by the National Research Foundation of Korea through the Ministry of Science and ICT under Grant 2021R1A2C1093957, by Korean Ministry of Land, Infrastructure, and Transport(MOLIT) as Innovative Talent Education Program for Smart City, by the Institute of Engineering Research at Seoul National University, and by the BK21 FOUR (Fostering Outstanding Universities for Research) funded by the Ministry of Education(MOE, Korea) and National Research Foundation of Korea(NRF).
\bibliographystyle{IEEEtran}
\bibliography{root}
\end{document}